\documentclass[12pt]{article}
\usepackage[top=1.0in, bottom=1.0in, left=0.75in, right=0.75in]{geometry}
\usepackage{amsmath, amsthm, amssymb}
\allowdisplaybreaks

\usepackage[affil-it]{authblk}
\usepackage{graphics}
\usepackage{mathrsfs}
\usepackage{setspace}
\usepackage{graphicx}
\graphicspath{ {./images/} }
\usepackage{epstopdf}
\usepackage{rotating}
\usepackage{longtable}
\usepackage{tabularx}
\usepackage{tabulary}
\usepackage{ltxtable}
\usepackage{multirow}
\usepackage{listings}

\newsavebox\affbox
\title{\textbf{Intelligent Systems Design for Malware Classification Under Adversarial Conditions}}

\author[1]{Sean M. Devine%
	\thanks{Email: \texttt{sean.devine@westpoint.edu}; Corresponding author}} 
\author[1,2]{Nathaniel D. Bastian, PhD}

\affil[1]{Department of Systems Engineering, U.S. Military Academy\\West Point, New York 10996}
\affil[2]{Army Cyber Institute, U.S. Military Academy\\West Point, New York 10996}

\date{16 May 2019}

\begin{document}
\maketitle

\abstract{The use of machine learning and intelligent systems has become an established practice in the realm of malware detection and cyber threat prevention. In an environment characterized by widespread accessibility and big data, the feasibility of malware classification without the use of artificial intelligence-based techniques has been diminished exponentially. Also characteristic of the contemporary realm of automated, intelligent malware detection is the threat of adversarial machine learning. Adversaries are looking to target the underlying data and/or algorithm responsible for the functionality of malware classification to map its behavior or corrupt its functionality. The ends of such adversaries are bypassing the cyber security measures and increasing malware effectiveness. The focus of this research is the design of an intelligent systems approach using machine learning that can accurately and robustly classify malware under adversarial conditions. Such an outcome ultimately relies on increased flexibility and adaptability to build a model robust enough to identify attacks on the underlying algorithm.\\

}

\newpage

\doublespacing

\section{Introduction}

Machine learning and intelligent systems seem to be at the forefront of the application of data science and analytics to the realm of cyber security and defense. While the contemporary cyber domain has become characterized by a reliance on big data and a widespread prevalence of threats, so has the need for increased automation with regard to analytics and data science techniques for active cyber defense. With regards to the classification of malware, intelligent systems offer a solution to the diminishing feasibility of data classification without automation. By assessing patterns within the data, machine learning classifiers are able to effectively label data as malicious, thus increasing the level of security and the ability of administrators to more effectively monitor their systems. Such classifiers rely on an existing pool of inputs and classifications in order to train the classifier. Mathematically, and given viable training data, machine learning classifiers are able to predict classification labels with a high degree of accuracy. Such a fact explains their applicability and effectiveness in classifying malware. In identifying data-based features of the malware, the classifier can effectively predict whether or not a given set of features correspond to an identified type of malware. 

From a cyber security perspective, these classifiers are extremely beneficial to the area of malware detection. However, the design of these classifiers is ultimately flawed and features key vulnerabilities that adversaries may exploit in order to diminish the classifier's effectiveness. Because these models rely on a pool of training data in order to shape its predictions, this data serves as a target for potential adversaries who may targeting the classifier's underlying algorithm. In perturbing, or poisoning, the training data that these classifier draw upon to shape its predictions, adversaries can lower the accuracy of these classifiers, or even alter the range of acceptable data, thus facilitating the obscuration of potential malware.

This research attempts to address this issue using an intelligent systems design approach by demonstrating a method of adding a degree of robustness to the design of classifiers in order to account for the potential data poisoning. This research analyzes the effect of purposefully training a series of base classification models on a variety of data perturbations to determine which models are best suited for optimal performance under given adversarial conditions. These models will then be combined through a linear stacking method in order to test the accuracy of classifiers that incorporate varying degrees of robustness.

\subsection{Literature Review}

The foundation of this research is in the field on machine learning and intelligent systems as they stand as the target of such adversarial machine learning attacks. As such, it is important to have an understanding of these machine learning systems within the realm of intelligent or expert systems. In defining the unique qualities of intelligent systems, Gregor and Benbasat (1999) layout the distinguishing factor of these systems as the presence of a ``knowledge component" or a computer representation of ``human tacit and explicit knowledge" \cite{GregorBenbasat1999}. In essence, such systems seek to use the ``knowledge component" as a means of automating computational tasks. F. Hayes-Roth (1997) offers a more in-depth analysis of basic elements of such systems that involve this artificial intelligence factor. He defines the basic elements of artificial as control, inference, and representation \cite{HayesRothF1997}. These elements compliment the earlier work of B. Hayes-Roth (1995) in which she defines the three basic functions of ``intelligent agents." B. Hayes-Roth defines these functions as a process of first generating a perception of the dynamic environment following by a corresponding action to affect such environmental conditions, and ending with an output based on interpreting the perceptions, drawing inferences, and determining appropriate actions \cite{HayesRothB1995}. 

In application of these principles, B. Hayes-Roth (1995) offers the concept ``adaptive intelligent systems" to combat the notion of designing systems to meet a specific niche. In contrast, B. Hayes-Roth proposes instead designing architectures that are fit to classes of niches and thus are much more adaptive the dynamic conditions within the environment \cite{HayesRothB1995}. Practically, intelligent systems serve to automate the processes of tasks such as pattern detection, cluster analysis, and intrusion detection \cite{Liu2018}. Machine learning systems have even become essential to big data analytics and data mining. However, its widespread use has led to the identification of various vulnerabilities and its identification by Yu (2016) as a major privacy concern \cite{Yu2016}. Such vulnerabilities are in part due to the emergence of the adversarial environment. Biggio et al. (2014) point the overall design of such systems as the major flaw. Essentially, the design of such systems are based on classical methods and evaluation techniques that do not consider the unique adversarial environment \cite{Biggio2014}.

The concept of the adversarial environment characterizes the current challenges facing machine learning based classification algorithms as they are subject to attacks to degrade their validity and effectiveness. Lowd and Meek (2005) describe this condition as they introduce the Adversarial Classifier Reverse Engineering (ACRE) problem which describes the adversary's approach to learning about a malware classifier \cite{LowdMeek2005}. Huang et al. would complement Lowd and Meek with their own application of these principles to their concept of adversarial machine learning. Huang et al. (2011) essentially apply a game theoretic approach to the adversarial learning problem, an approach in which the adversary is actively attempting to determine strategy of the classifier in order to increase the effectiveness of malware attacks \cite{Huang2011}. They also reference two forms of attacks within adversarial learning. The first form is the exploratory attack in which the adversary essentially observes the classifiers response to tailored instances. The second form is the causative attack in which the adversary attempts to corrupt the training data in order to influence a false mapping and thus produce a poor classifier \cite{Huang2011}. 

Khurana, Mittal, and Joshi (2018) offer an analysis of such attacks in the contemporary adversarial learning environment as they describe an approach to combating poisoning attacks. Such attacks involve the poison of open-source intelligence (OSINT) data sources with instances designed to produce false positives/negatives by threat defense ar) systems \cite{Khurana2018}. Their approach consisted of modeling the credibility of OSINT prior to its incorporation into threat detection AI training data \cite{Khurana2018}. The concept of the poisoning attack is more specifically the subject of the online centroid detection. According to Kloft and Laskov (2012), poisoning attacks attempts to target the centroid and radius of the area of classifier’s range of legitimate classifications. With the introduction of malicious training points, an adversary can force the centroid to shift in the direction of the attack. Doing so can allow new attacks to fall within the range of what the classifier considers legitimate data \cite{Kloft2012}.

The existence of these adversarial capabilities, as well as this initial understanding of the adversarial model presents the concept of the arms race within adversarial machine learning. Biggio et al. (2014) present this idea as a means of understanding the relationship between classifier and adversary \cite{Biggio2014}. The arms race is separated between the reactive arms race, in which a classifier develops counter measures following an attack, and the proactive arms race in which classifiers attempt to anticipate the adversary’s actions based on the understanding of the adversarial model \cite{Biggio2014}. Berreno et al. (2006) define the adversarial model with their concept of the attack model. The attack model looks to classify adversarial attacks based on influence, specificity, and security violation of the attack. Within the influence, the model defines the attack as wither causative or exploratory. Specificity expands the classification by identifying the attack as either a targeted attack on a specific set or point, or indiscriminate. The final level of classification is the form of security violation. An integrity attack looks to influence the classifier to produce false negatives. An availability attack is broader as it looks to produce both false positives and false negatives \cite{Barreno2006}. Such a classification of an attack aids in understanding what Biggio et al. (2014) refer to as the overall adversarial goal, knowledge, capabilities, and strategy \cite{Biggio2014}.
	
Inherent to the proactive arms race is understanding the implications the concept has on classifier design considerations. Within the design of such system, Papernot et al. (2016) introduces the concept of the no free lunch theory. The major benefit to machine learning based intelligent systems is the inherent simplicity of their hypotheses. While increasing the robustness of such hypothesis would decrease their inherent vulnerabilities, doing so comes at the price of the accuracy of the system in correctly classifying legitimate inputs \cite{Papernot2016}.

\subsection{Research Objectives} 

The central objective of this research is to respond to the need for a malware classification model that is robust enough to operate effectively within the emerging adversarial environment. While the development of such models has been the subject of established research, the new difficulty in this field is in testing and assessing such models for robustness. The underlying concern is thus the development for processes of experimentally testing classification models. Involved in this process is the critical task of identifying and utilizing known classification performance metrics and potentially new metrics or measures of effectiveness.

The end result of this research is ultimately a model robust enough to accurately classify malware within an adversarial environment. Increased model robustness will decrease the model's susceptibility to poisoning attacks. Such attacks are the mechanism behind which adversaries alter a machine learning algorithm's range of acceptability. In achieving this end result, the notable secondary result is an effective process by which models can be experimentally tested and assessed for effectiveness. Such a product could prove to be useful for future machine learning model development as intelligent design techniques undergo their own development to meet the demands of the adversarial environment.  

\section{Materials and Methods}

This research follows the Cross Industry Standard Process for Data Mining (CRISP-DM) methodology, the stages of which can be seen in Figure \ref{figure:1}. The first three phases of the methodology (Business Understanding, Data Understanding, Data Preparation) encompass the overall initial preparation and foundation for modeling and evaluation. The final three phases of Modeling, Evaluation, and Deployment occur as part of the general approach to the development and experimental testing and assessment of AI-based malware classifiers. for more details about the CRISP-DM, please refer to \cite{crisp2000}.
\begin{figure}[h!]
\includegraphics[width=8.5cm]{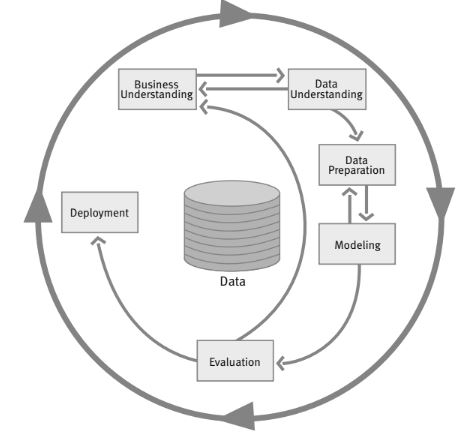}
\centering
\caption{Stages of the CRISP-DM Process Model \cite{crisp2000}}
\label{figure:1}
\end{figure}

In an effort to incorporate robustness into the intelligent systems design of malware classifiers, this research relies on stacking as a means of building robustness into a supervised learning-based classification model. This approach, as applied to this research, begins with a comparison of the accuracy of 10 base classification models when trained on three data sets of varying perturbation, or ``poisoning", schemes. This comparison ultimately reveals the most accurate model per data set. In stacking the top performing models in terms of accuracy, this research looks to reveal that a classification model can be developed to account for potential forms of training data perturbation with minimal effect on the overall model accuracy.  

\subsection{Data Understanding}

The modeling for this research is based on a training data set of samples previously identified as malware. Particularly, the samples were identified by a class corresponding to its malware type. The set of malware types used for this dat aset consist of Viruses, Worms, Packed Malware, and AdWare. The data set is in the form of a $N$x$M$ matrix in JSON format and consists of six schema: ``data", ``row\_index",``column\_index",``schema",``shape", and ``labels." More specifically, the schema take on the following ranges and data types:

\begin{center}
     data : int [1:6317199]\\
     row\_index   : int [1:6317199]\\
     column\_index: int [1:6317199]\\
     schema : chr [1:106428]\\ 
     shape : int [1:2]\\
     labels : int [1:12536]
\end{center}

Among the schema present in the JSON file, ``data", ``row\_index", and ``column\_index" correspond to the contents of the feature matrix used to train potential classifiers. The feature matrix represents the instance, \(i\), that corresponds to the input, 
\([row\_index[i], column\_index[i]]\). As such, a particular instance in the ``data" schema is represented by the following function:
\begin{equation}
    feature\_matrix[row\_index[i], column\_index[i]] = data[i]
\end{equation}
The contents of ``schema" take on the form of either unigrams, bigrams, or trigrams that correspond to a particular Windows API call. The ``shape" schema corresponds to the \([N,M]\) values that dictate the dimensions of the data matrix. Finally, the ``label" scheme correspond to the class of each instance within the feature matrix. This schema is unique to the training data as it is used as a means of assessing the accuracy of classification models.

\subsection{Data Preparation}

Preparing the data for analysis and model input involved first iterating through the range of the ``data" schema according to the function defined in (1), At this point the feature matrix can then be converted to compressed sparse row (CSR) format and saved to a npz file. Furthermore, this initial CSR format of the data can be normalized with scaling from -1 to 1 and a fixed standard deviation.

While the described technique provides a feature matrix that can be used to train classification models, the shape of this initial matrix (12536, 6317199) poses an issue of dimensionality. High-dimensional data in general is an issue based on in increased need in the number of samples necessary for a estimator to effectively generalize. Furthermore, in reducing the dimensions, the data can be analyzed and used for training classifiers despite limitations to memory and processing power. For the purpose of this research, responding to the problem of high-hdimensionality involved an implementation of singular value decomposition (SVD) on the feature matrix. This SVD technique is the equivalent to Principle Components Analysis in which the columnwise mean is subtracted from the feature values. Such a technique can be tailored so as to produce new, reduced data sets based on a desired number of dimensions. Given the initially prepared data set, the feature matrix can be reduced to 1000, 500, 250, and 100 dimensions respectively. Similar to the initial data set, these reduced data sets are formatted as CSR data sets and written to npz files. The approach is implemented using the \textit{truncatedSVD} class with \textit{sklearn} in Python 3.7.

For the purpose of training the classifiers, the data sets generated thus far represent the input (features). The output, or prediction, of such classifiers is ultimately the classification label (0-4), a categorical variable. Such labels can be extracted from the initial JSON file of the training data. Analysis and comparison of the classification models in this research is based on the results of the models when trained on both an unperturbed, base data set, and when trained on data sets containing a perturbed set of inputs and a perturbed set of labels, respectively. More specifically, the classification models are each trained on three separate data sets and are assessed based on a constant validation data set containing inputs and the corresponding labels that are pulled from the initial base data set. 

The validation set is generated by designating 25\% of the base data set, \(Tr_{input}\) as a constant, unaltered data set with the naming convention \(V_{input}\). The labels that correspond to the input are similarly designated as \(V_{label}\). These data sets serve as method of comparing each classifier training iteration (regardless of input) against an immutable validation set. The remaining 75\% of the base data set is used as the unperturbed control data set, which is designated as \(C_{1, input}\) with the
corresponding set of labels being designated as \(C_{1, label}\). As the \(C_{1, input}\) and \(C_{1, label}\) serve as the unperturbed training set, they act as the control training set to determine the relative effect of the perturbed training sets on each malware classifier.

The overall perturbation scheme entails targeting the two aspects of training data that serve as potential threat vectors for adversaries. In an attempt to poison the data, an adversary could target either the input features themselves, or the corresponding training labels in order to degrade the classification accuracy. The first set of perturbations on the base data set aimed to naively emulate the poisoning of the data inputs (features). The scheme is based on applying a uniform perturbation to a percentage of the features per input so as to effectively poison the data and alter the accuracy of a potential classification, without being so drastic as to easily recognizable. The perturbation is based on iterating through the base inputs of \(C_{1, input}\) to randomly perturb 20\% of the features per input. Of the values selected, the perturbation involves multiplying non-zero values by 1.5 and changing zero values to the product of a random integer between 1 and 10 and 0.1. This perturbed data set is designated as \(C_{2, input}\). The set of corresponding labels, \(C_{2, label}\), is unperturbed and, therefore, is equal to \(C_{1, input}\).

The second set of perturbations on the base data set aimed to naively emulate the poisoning of training data classification labels. Given the lower volume of data points in the training labels data set compared to the training inputs data set, the goal with this scheme was to more randomly perturb a percentage of the labels in order to degrade the classification accuracy. Therefore, this perturbation scheme is based on iterating through the base classification labels of \(C_{1, label}\) to randomly perturb 20\% of the labels within the base data set. The values selected are changed to a random value between 0 and 4, representative of the different potential malware types. This data set is designated as \(C_{3, label}\). Meanwhile, the set of corresponding inputs is unperturbed and, therefore, is equal to the inputs in data set \(C_{1, input}\).

\subsection{Model Building}

The foundation of this research is the implementation of the 10 most commonly used machine learning models for supervised learning (classification): random forest, support vector machine, gradient boosting, logistic regression, artificial neural network, linear discriminant analysis, quadratic discriminant analysis, naive bayes, bagging, and decision tree. All models are implemented using the Python 3.7 machine learning library, scikit-learn, or \textit{sklearn}, which facilitates simplicity and efficiency in machine learning for data mining and analysis. For the sake of this research, all models are implemented with their respective default parameters; hence, no hyper-parameter tuning was performed, serving as future work.

\begin{figure}[h!]
\includegraphics[width=9cm]{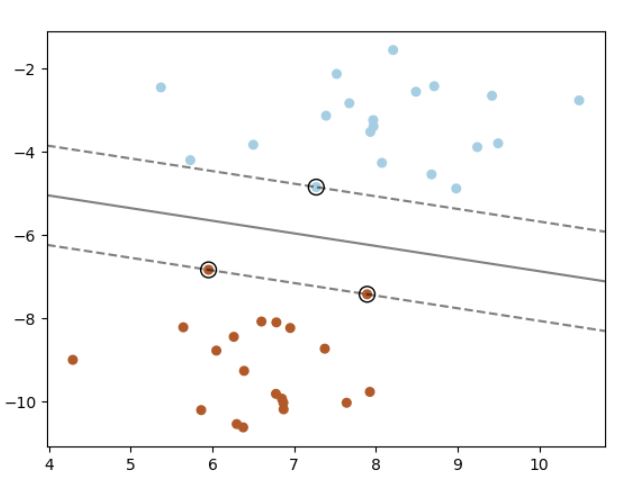}
\centering
\caption{Optimal Functional Margin Between Two Points of Two Respective Classes \cite{sklearn}}
\label{figure:2}
\end{figure}

The support vector machine classifier calculates a set of hyperplanes based on the training data, all of which would potentially classify the inputs. In order to identify the optimal hyperplane, the classifier then looks to maximize the functional margin as depicted in Figure \ref{figure:2}. The margin ultimately represents the distance between the plane and any point of either class. The classifier is implemented using the \textit{svm} class within \textit{sklearn}.
As applied to multi-class classification, the logistic regression classifier distinguishes between classes. The classifier essentially follows a multinomial logistic function in order to model a response variable that describes the probability prediction of the classification label \cite{Hackling}.

\begin{figure}[h!]
\includegraphics[width=10cm]{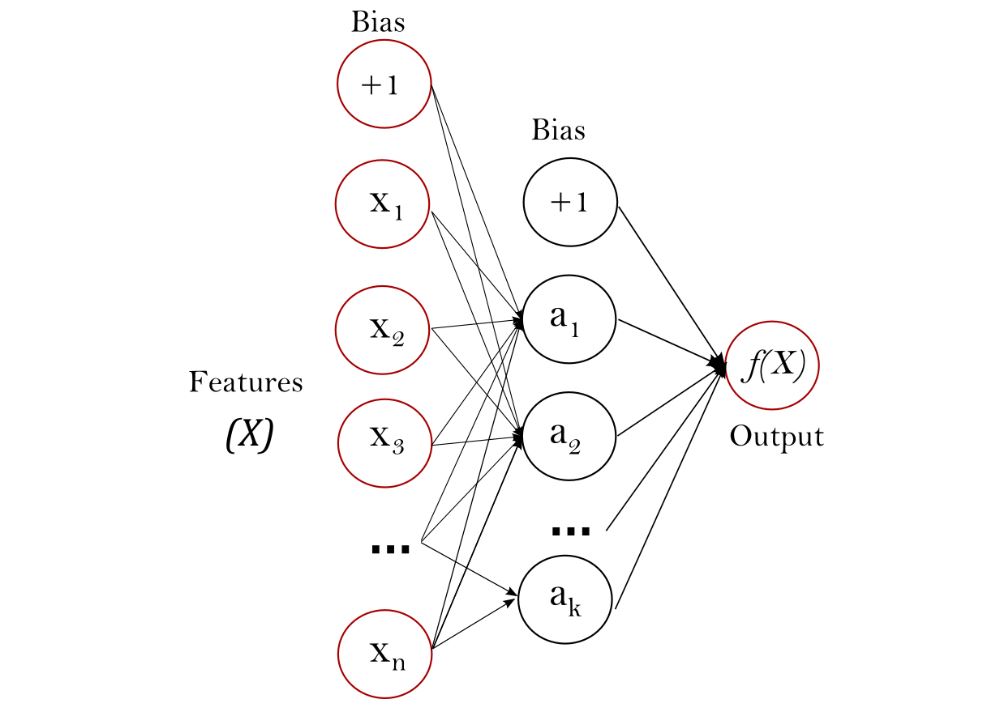}
\centering
\caption{Feed-Forward Artificial Neural Network Containing a Hidden Layer of Perceptrons \cite{sklearn}}
\label{figure:3}
\end{figure}

The multi-layer perceptron classifier, as depicted in Figure \ref{figure:3}, is an implementation of an artificial neural network. It is a base implementation of the \textit{MLPClassifier} class in \textit{sklearn}. The artificial neural network consists of hidden layers of perceptrons. Within each layer, inputs are transformed based on trained weight summation, or bias. Weights at each hidden layer are trained using backpropagation, relying on gradient to minimize a cross-entropy cost function. The result is ultimately a set of prediction probability estimates for each input \cite{sklearn}.

Linear discriminant analysis (LDA) and quadratic discriminant analysis (QDA) are based on training linear and quadratic decision boundaries within each respective model. Decision boundaries for both models are generating according to Bayes' rule, with the conditional probability being modeled on a Gaussian distribution. the difference between the two models is in the assumed covariances of the Gaussian densities for each class. 
\begin{figure}[h!]
\includegraphics[width=13cm]{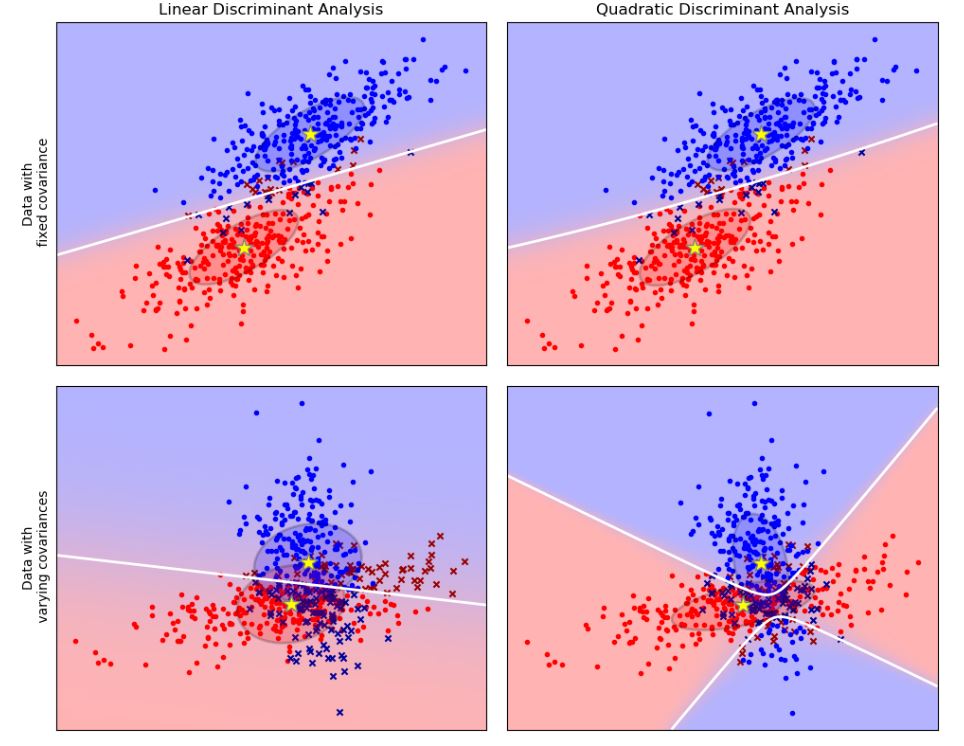}
\centering
\caption{Differences Between LDA and QDA Given Data with Fixed and Varying Covariances \cite{sklearn}}
\label{figure:4}
\end{figure}
In LDA, the decision boundaries are calculated under the assumption that the densities for each class share the same covariance matrix. In QDA, there are no assumptions with regard to the covariance matrix for each density. The result, therefore, is the possibility for quadratic decision boundaries as opposed to the strictly linear decision boundaries for LDA \cite{sklearn}. The effect of the differing treatment of covariances can be more definitively depicted in Figure \ref{figure:4}. Figure \ref{figure:4} especially highlights how this differing treatment can effect the decision boundaries for the LDA and QDA classifiers respectively, as the QDA will produce a quadratic decision boundary when trained on data with varying covariances. The two models are implemented using the \textit{sklearn} classes \textit{LinearDiscriminantAnalysis} and \textit{QuadraticDiscriminantAnalysis}, respectively.

The naive bayes classifier is an application of gaussian naive bayes classification. The model is an inherently a naive classification approach due to its assumption of conditional independence between pairs of features for a given input. This particular approach assumes the value of the likelihood of the features follows the gaussian distribution in order to calculate its prediction probabilities \cite{sklearn}.

The decision tree classifier is essentially a non-parametric approach to classification that model a decision, or classification, as tree-like graphs. The ultimate classification is based on the linear combination of explanatory variables for a given input. The model seeks the shortest sequence of explanatory variables in order to calculate its prediction \cite{Hackling}. Developing the model calls for a recursive partitioning of a training sets with respect to its explanatory variable, so that inputs with similar features are grouped together. The recursive partitioning creates the structure of a tree and facilitates such analysis of the shortest combination of features that enables a prediction \cite{sklearn}. In this research, the model is implemented through the \textit{DecisionTreeClassifier} class within \textit{sklearn}.

The random forest classifier expands upon the decision tree in that it is an ensemble method consisting of a series of trees derived from randomized subsets of the training data. Similar to the base decision tree, each tree will produce a probabilistic prediction of the overall classification. This prediction probability is then averaged across all trees, thus producing an overall prediction for the given input features \cite{sklearn}. The model is implemented using the \textit{RandomForestClassifier} and this implementation consists of 100 trees derived from the corresponding number of randomized samples from the training data. The bagging classifier is another ensemble method and for this research is implemented using a very similar approach. The key differences in the random subset of the training data, features are drawn without replacement. The bagging classifier is implemented suing the \textit{BaggingClassifier} class \cite{sklearn}.

In a similar approach, the gradient boosting classifier is an ensemble method that relies on the use of a series of decision trees as weak learners. The classifier then builds a greedy, additive model in which trees are added so as to minimize loss. The minimization is achieved by calculating the steepest descent method. The method is ultimate a numerical approach and involves a calculation of the gradient of the loss function for the given stage of the ensemble \cite{sklearn}. The base model is implemented using the \textit{GradientBoostingClassifier} class within \textit{sklearn}.

Finally, this research utilizes a stacking method in an attempt to generate a more robust model. The linear stacking method involves designating models to be stacked and such stack serves as Level 1 of the model. In ``stacking" the approach. as depicted in figure \ref{figure:1}, trains and runs the selected models in order to capture each model's respective predictions. The predictions for each are then compiled to form a new data set in which the predictions become the features for each input and correspond to a label in the validation or test set. The new ``stacked" data set is then pushed to Level 2 of the linear stacking model which consists of a base classifier. The Level 2 classifier is trained on the new data set so that it is essentially predicting a label based on the predictions of the Level 1 models \cite{stacking}.
\begin{figure}[h!]
\includegraphics[width=12cm]{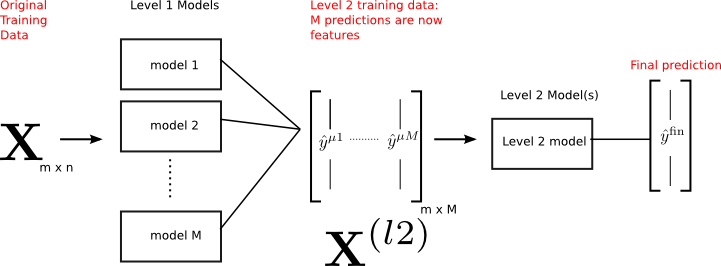}
\centering
\caption{Description of the Data Flow Within the Linear Stacking Method}
\label{figure:1}
\end{figure}

\section{Results and Discussion}

\subsection{Base Model Deployment}

The base models serve as the foundation of this research, as their classification accuracy when trained on \(C_{1}\),\(C_{2}\), and \(C_{3}\), respectively, determines the Level 1 make up of stacked across the top performing models per data set. Table \ref{table:1} list the results of deploying the base models. Each model was trained on each of the three data sets, and the accuracy report is based on comparing the model prediction to the \(V_{label}\) set.

\begin{table}[h!]
\caption{Accuracy Scores for Base Models based on Given Input Data Sets \(C_{1}\), \(C_{2}\), and \(C_{3}\)}
\label{table:1}
\centering
\begin{tabular}{ |p{4.5cm}||p{1.25cm}|p{1.25cm}|p{1.25cm}| }
\hline
\multicolumn{4}{|c|}{\textbf{Base Model Classification Accuracy}} \\ [1ex]
\hline\hline
\textbf{Base Classifier} &  \(C_{1}\)    & \(C_{2}\) & \(C_{3}\) \\ [0.5ex]
\hline 
Random Forest & 0.9294 & 0.9262 & 0.8931 \\

Support Vector Machine & 0.9135 & 0.9007 & 0.9103 \\

Gradient Boosting & 0.9211 & 0.9175 & 0.9111 \\

Logistic Regression & 0.9007 & 0.8923 & 0.8927 \\

Artificial Neural Network & 0.9167 & 0.9175 & 0.8927 \\

LDA & 0.8836 & 0.8820 & 0.8880 \\

QDA & 0.8050 & 0.6886 & 0.5060 \\

Naive Bayes & 0.5602 & 0.4944 & 0.5817 \\

Bagging & 0.9226 & 0.9215 & 0.8668 \\

Decision Tree & 0.8919 & 0.8967 & 0.8002 \\ [1ex]
\hline 
\end{tabular}
\end{table}

Initial analysis of the classification model accuracy reveals the general trend of data set \(C_{2}\), a perturbation of the features, causing a slight drop in the overall accuracy of the models and \(C_{3}\), a perturbation of the labels, causing a more drastic decrease in accuracy. The main take away from the base model deployment, however, is that given the respective perturbation schemes, the results in Table \ref{table:1} provide insight into the top performing models, and, thus, the models most likely to add the desired level of robustness to the stacked model.

For data set \(C_{1}\), the random forest classifier is the best performing model based on its accuracy of 0.9294. It is key to note, however, that while not the top performing models relative to accuracy, the bagging and gradient boosting classifiers produce similarly high accuracies of 0.9226 and 0.9211, respectively. In the case of the bagging classifier, it appears to be the second highest performing classifier for both data sets \(C_{1}\) and \(C_{2}\), respectively. In both deployments, the model is outperformed by the random forest classifier. Given that the random forest classifier is the top performer for data set \(C_{1}\), a second iteration of the model should not be included in the stacked model as doing so would only add duplicate label predictions from the same source. Because of this, this research designates the bagging classifier as the top performer for data set \(C_{2}\). The bagging classifier's performance when trained on \(C_{1}\) also adds a degree of relative surety that it is a strong candidate for inclusion in the stacked model.

Analysis of the gradient boosting classifier demonstrates that although it is not a top performer for data set \(C_{1}\) or data set \(C_{2}\), its accuracy of 0.9111 when trained on data set \(C_{3}\) is the highest among all of the base classifiers. It also is key to note that while the perturbation scheme in data set \(C_{3}\) seems to have had a relatively drastic effect on the other base models, its effect on the accuracy of the gradient boosting classifier is relatively minimal.

\subsection{Stacked Model Development and Deployment}

The stacked model draws upon these top performing models across the three data sets in order to generate the Level 1 classification. Based on each classifier's predicted label for each input, the Level 2 classifier in the stack will predict an overall label for a given input in the data set. Given the Level 1 stack generated by the random forest, bagging, and gradient boosting classifiers and the original 10 base classification models, the product is ultimately 10 stacked models all based on a constant Level 1 stack. Similar to the base models, these 10 stacked models can be trained on \(C_{1}\), \(C_{2}\), and \(C_{3}\) respectively in order to compare model accuracy relative to each perturbation scheme. Table \ref{table:2} lists the accuracy of each model relative to each data set as a means of comparing model performance. 

Ultimately the key takeaway from the results of the stacked model is their performance relative to the Table \ref{table:1} results. When trained on \(C_{1}\) the random forest classifier predicts with an accuracy of 0.9294, when trained on \(C_{2}\) the bagging classifier predicts with an accuracy of 0.9215, and when trained on \(C_{3}\) the gradient boosting classifier predicts with an accuracy of 0.9111. In analyzing the Table \ref{table:2} results, it is clear that the stacked models do not outperform the top performing base model relative to data sets \(C_{1}\), \(C_{2}\), and \(C_{3}\), respectively. However, it is key that the stacked models achieve the goal of increased robustness, with only a minimal effect on the accuracy of the classifier. Specifically, the stacked model with a Level 2 support vector machine classifier achieves an accuracy of 0.9183, 0.9147, and 0.8900 for data sets \(C_{1}\), \(C_{2}\), and \(C_{3}\), respectively.

\begin{table}[h!]
\caption{Accuracy for ``Best of" Stacked Models Based on Given Input Data Sets \(C_{1}\), \(C_{2}\), and \(C_{3}\)}
\label{table:2}
\centering
\begin{tabular}{ |p{4.5cm}||p{4.5cm}||p{1.25cm}|p{1.25cm}|p{1.25cm}|} 
\hline
\multicolumn{5}{|c|}{\textbf{Stacked Model Classification Accuracy}} \\ [1ex]
\hline\hline
\textbf{Level 1 Stack} & \textbf{Level 2 Classifier} & \(C_{1}\) & \(C_{2}\) & \(C_{3}\) \\ [0.5ex]
\hline
\multirow{11}{4em}{Random Forest, Bagging, Gradient Boosting} & Random Forest & 0.9091 & 0.9095 & 0.8768\\ 
&Support Vector Machine & 0.9183 & 0.9147 & 0.8900 \\ 
&Gradient Boosting & 0.9099 & 0.9099 & 0.8856 \\ 
&Logistic Regression & 0.6800 & 0.8297& 0.6826 \\ 
&Artificial Neural Network & 0.9163 & 0.9115& 0.8808 \\ 
&LDA & 0.7041 & 0.7085& 0.6882 \\ 
&QDA & 0.8696 & 0.8788& 0.8589 \\ 
&Naive Bayes & 0.9167 & 0.9135& 0.8844 \\ 
&Bagging & 0.9111 & 0.9063& 0.8824 \\ 
&Decision Tree & 0.9111 & 0.9091 & 0.8864 \\ [1ex]
\hline
\end{tabular}
\end{table}

In effort to analyze the effects of adding a further degree of robustness to the model design, this research considers two models based on a Level 1 stack of all base models with a Level 2 consisting of the gradient boosting classifier and the support vector machine classifier, respectively. In analyzing the accuracy of these two additional stacked models, which is provided in Table \ref{table:3}, it is interesting to note that not only does this approach increase the level of robustness, the models also outperform the stacked models based on the top performing base models. The stack of all models with the support vector machine as the Level 2 classifier even outperforms the top performing base models for data sets \(C_{2}\) and \(C_{3}\).

\begin{table}[h!]
\caption{Accuracy for Stack of All Base Models Based on Given Input Data Sets \(C_{1}\), \(C_{2}\), and \(C_{3}\)}
\label{table:3}
\centering
\begin{tabular}{ |p{4.5cm}||p{1.25cm}|p{1.25cm}|p{1.25cm}| }
\hline
\multicolumn{4}{|c|}{\textbf{Classification Accuracy of a Stack Across All Models}} \\ [1ex]
\hline\hline
\textbf{Level 2 Classifier }&  \(C_{1}\)    & \(C_{2}\) & \(C_{3}\) \\ [0.5ex]
\hline 
Gradient Boosting & 0.9222 & 0.9222 & 0.9091 \\ 
Support Vector Machine &0.9274&0.9264&0.9179\\[1ex]

\hline 
\end{tabular}
\end{table}

Given this overall top performing model, a recommendation for subsequent implementation would be to further expand on this intelligent systems approach with another ensemble method; this would further increase the overall robustness of the model. Given that the stack of all models with the Level 2 support vector machine model produced a prediction based on each of the three data sets, \(C_{1}\), \(C_{2}\), and \(C_{3}\), the three models can be linearly combined relative to their predicted probabilities. Given the three sets of predicted probabilities per class per input, the maximum a posterior sum of probability values per class per input can be used to determine the overall prediction per input. Specifically, the result will be a single set of vectors corresponding to each input of the base data set. Within each vector, select the highest value, just as the highest prediction probability is selected to produce a classification prediction; this is also known as a soft voting ensemble. The ultimate result is a robust classification that is designed intelligently to account for the potential for a base, unperturbed data set, a perturbation of features, and a perturbation of labels. 

\section{Conclusions and Future Work}

\subsection{Conclusions}

This research using an intelligent systems design approach for malware classification ultimately called for developing and training machine learning models based on a purposeful poisoning, or perturbation, of the training data. This research ultimately looked to demonstrate that doing so aided in identifying those base models best equipped to consider varying degrees of data poisoning. Stacking these models created a classifier that is designed with robustness in mind and considers the potential of poisoned training data, all while limiting that effect of poisoning and increased robustness on the model's classification accuracy.

The results of this research provide evidence that this method of stacking as a means of accounting for data poisoning increased robustness without a detrimental effect on the overall accuracy of the model. Furthermore, in expanding the robustness by increasing the number of models in the Level 1 stack, the stacked model is better equipped to classify in adversarial conditions than the base models alone. The trade off of is a slightly diminished classification accuracy in instances where training data may be unperturbed.

Overall, the Level 1 stack across all models with a Level 2 support vector machine can be considered the best performing model, as it outperformed even the base models when considering adversarial conditions and perturbations within training data sets. There of course exists even more potential for this model to be expanded using a final soft voting ensemble approach designed to account for all perturbation schemes at once. This implementation can only further expand the robust, intelligent systems design of the malware classification model, making it more apt to handle the potential for adversarial conditions.

\subsection{Limitations and Future Work}

First, the shear size of the data set and the number of features per input exposed a significant memory limitation of the machine used for classification model design and implementation. This memory limitation drove the need for a high degree of dimensionality reduction of the initial base data set. The effect was minimal due to the scope and objective of this research. Future work, however, may call for analysis of the models on larger, non-dimension reduced data sets. 

Second, the data poisoning schemes used within the intelligent systems design approach incorporated relatively naive perturbations based upon simple stochastic sampling. As a result, future work should investigate and compare more sophisticated data poisoning schemes to conduct the perturbations, such as genetic algorithms, particle swarm optimization, generative adversarial networks, or other heuristic-based methods.

Third, future work calls for the optimization of the initial base classification models. Given the objective of this research, using the ``default" status of the models was sufficient due to the focus on developing a robust, intelligent systems design approach. However, future efforts should focus on refining the base models prior to the stacking efforts. In particular, a full- or partial-experimental design could be used for hyper-parameter tuning of each of the malware classifiers. By optimizing the hyper-parameters of the base models, this could prove to optimize the classification accuracy of the final stacked model while maintaining its inherent robustness.

Finally, future work should test and operationalize the final soft voting ensemble approach to add the prediction probabilities of the final stacked model for each data set. This research provides somewhat of an experimental foundation, but further research should focus on practical implementation of the intelligent systems design within malware classification.

\end{document}